\documentclass{article}
\usepackage{ijcai25}

\usepackage{times}
\usepackage{soul}
\usepackage{url}
\usepackage[hidelinks]{hyperref}
\usepackage[utf8]{inputenc}
\usepackage[small]{caption}
\usepackage{graphicx}
\usepackage{amsmath}
\usepackage{amsthm}
\usepackage{booktabs}
\usepackage{algorithm}
\usepackage{algorithmic}
\usepackage[switch]{lineno}
\usepackage{natbib}
\usepackage[table]{xcolor}    
\usepackage{array}     
\usepackage{multirow}
\usepackage{multicol}

\newcommand*\colourcheck[1]{%
  \expandafter\newcommand\csname #1check\endcsname{\textcolor{#1}{\ding{52}}}%
}
\colourcheck{blue}
\colourcheck{green}
\colourcheck{red}
\usepackage{pifont}
\newcommand{\xmark}{\textcolor{red}{\ding{55}}}%

\title{A Comprehensive Survey on the Risks and Limitations of Concept-based Models}

\author{
Sanchit Sinha$^1$
\And
Aidong Zhang$^1$\\
\affiliations
$^1$University of Virginia\\
\emails
\{sanchit, aidong\}@virginia.edu
}

\begin{document}

\maketitle

\begin{abstract}
Concept-based Models are a class of \textit{inherently explainable} networks that improve upon standard Deep Neural Networks by providing a rationale behind their predictions using human-understandable `concepts'. With these models being highly successful in critical applications like medical diagnosis and financial risk prediction, there is a natural push toward their wider adoption in sensitive domains to instill greater trust among diverse stakeholders. However, recent research has uncovered significant limitations in the structure of such networks, their training procedure, underlying assumptions, and their susceptibility to adversarial vulnerabilities. In particular, issues such as concept leakage, entangled representations, and limited robustness to perturbations pose challenges to their reliability and generalization. Additionally, the effectiveness of human interventions in these models remains an open question, raising concerns about their real-world applicability. In this paper, we provide a comprehensive survey on the risks and limitations associated with Concept-based Models. In particular, we focus on aggregating commonly encountered challenges and the architecture choices mitigating these challenges for Supervised and Unsupervised paradigms. We also examine recent advances in improving their reliability and discuss open problems and promising avenues of future research in this domain. 
\end{abstract}

\section{Introduction}
With the rising widespread adoption of Deep Neural Networks (DNNs) in human endeavors, there is a strong demand for transparency and rationales behind their decisions. This has resulted in significant research on the explainability of trained models as well as designing inherently explainable models for sensitive applications in medicine and finance. Early explainability research focused on providing feature attributions \cite{sundararajan2017axiomatic, ribeiro2016model} or sample importance \cite{koh2017understanding} - providing explanations local to specific features and data samples. However, for a more global explanation, explainability using `concepts' is overwhelmingly preferred. \textbf{Concepts} are abstract human-understandable entities, shared among multiple data samples and are directly relevant to task prediction. Multiple research works have attempted to discover concepts in trained models in a post-hoc manner \cite{kim2018interpretability, ghorbani2019towards} with various levels of success. These methods usually employ a set of pre-curated representative concepts as probes to query the models for each sample to be explained.

More recently, inherently explainable concept-based models have been at the forefront of explainability research. The process of utilizing concepts as intermediaries in task predictions is commonly called as \textit{concept learning}. Initial research lines of work have attempted to design architectures that assign importance scores to concepts defined in the input space such as \cite{ghorbani2019towards,chen2019explaining} and utilize them during predictions. However, such models suffer from sampling bias in the training and are limited to visually grounded, non-semantic concepts. More recent research attempts to learn concepts as an intermediate to task prediction, with concepts directly responsible for task prediction. In this paper, we focus on concepts defined in the representation space of models and not in the input space. Concept learning is broadly divided into categories of supervised -i.e. supervised with human-annotated concepts such as \cite{koh2020concept, espinosa2024learning} or unsupervised -i.e. learned automatically as soft signals from task prediction \cite{alvarez2018towards} (For a more detailed discussion refer Section~\ref{sec:paradigm}).

In this paper, we provide a comprehensive survey on the common risks and limitations faced during designing and deployment of concept-based models. We thoroughly explore and document common vulnerabilities in both supervised and unsupervised concept-based models. Even though a few surveys exist in the field of explainability, extremely few of them focus explicitly on concept-based explainability which is an intense and fast-growing field of research. To our knowledge, only the surveys by \cite{poeta2023concept} and \cite{shams2024navigating} focus on concept-based explainability but list only the types of architectures and do not document their vulnerabilities in detail. Elsewhere the survey by \cite{nannini2025nullius} focuses on the risks of Explainable AI but does not focus on Concept-based Models. We believe that a thorough analysis and discussion on the topic of risks and limitations encountered in concept-based models is highly beneficial in helping stakeholders design and deploy them in practice. In particular, our contributions in this paper can be summarized as:
\begin{itemize}
    \item We provide a comprehensive survey on the risks, limitations and vulnerabilities of Concept-Based Models, across both supervised and unsupervised paradigms.
    \item We also thoroughly document mitigation strategies for common vulnerabilities like concept leakage, lack of concept independence, adversarial vulnerabilities, intervention difficulties, etc.
    \item We provide guidelines for open research problems, and propose future directions to enhance the interpretability, robustness, and reliability of concept-based models in real-world applications.
\end{itemize}

\section{Paradigms of Concept Learning}
\label{sec:paradigm}
Inherently explainable concept-based models can be broadly divided into two distinct paradigms based on the availability of concept-level human annotations.

\noindent \textbf{Supervised Concept Learning:} The first paradigm treats the concept learning and task learning as independent problems with the concepts acting as a precursor to task prediction. As a result, it requires a set of human-annotated concept labels in addition to the task labels and treats concepts as \textit{signals} from concept learning. The concept labels are utilized as supervision for intermediate outputs during model training in addition to task labels. The requirement of concept-level supervision makes this approach \textbf{completely supervised} in the concept space. The seminal work in this field is Concept Bottleneck Models \cite{koh2020concept}, which demonstrated for the first time that \textit{any} DNN-based classification backbone can be converted into an equivalent inherently explainable concept-based model. In addition, the model supports \textit{intervention}. i.e. `correcting' a model's wrong prediction by intervening manually in the concept space. As human-annotated concept labels completely supervise the concept space, there is significant trust in model predictions. However, these approaches offer limited scalability due to the enormous human cost of manually annotating each concept label. Some newer approaches have utilized LLMs as knowledge banks to annotate concepts, reducing the human effort in labeling. We discuss them in Section~\ref{sec:future} as Large Language Models (LLMs) reduce human effort in labeling but still lie in the broad umbrella of supervised concept learning frameworks. 

\noindent \textbf{Unupervised Concept Learning:} The second paradigm attempts to learn concepts directly from the task itself without the presence of \textit{any} concept labels. This approach treats concepts as model \textit{representations} during model learning rather than explicit signals. The task labels act as weak supervision signals to guide concept learning and the lack of concept supervision makes this approach \textbf{completely unsupervised} in the concept space. As opposed to supervised approaches, these approaches do not provide explicit concept interpretation (like labels), which are left to stakeholders to figure out. The seminal work in this field is Self-Explaining Neural Network \cite{alvarez2018towards} which proposes a Siamese network architecture, wherein the first network extracts representative concepts while the second network learns relevance scores associated with the concepts. The final task prediction is performed as a weighted combination of the concepts and the associated scores. These approaches are seamlessly scalable, requiring no human effort in supervision. On the other hand, as there is no control over the learned concepts, these approaches are by definition low-trust and less reliable overall.

\begin{figure}[t]
    \centering
    \includegraphics[width=0.5\textwidth]{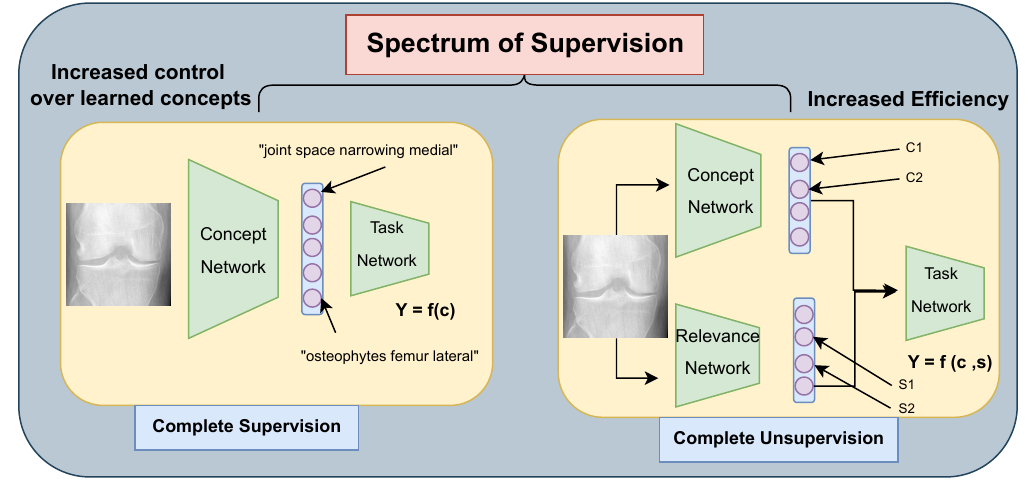}
    \caption{The Spectrum of Supervision in Concept Learning: An example taken from \cite{koh2020concept} wherein the task is predicting the KLG grade for osteoporosis of an X-ray of a knee. The concepts to look for include `joint space', and `osteophytes femur' - clinically sound concepts observed by doctors to adjudge the severity of osteoporosis. Complete supervision enables precise control over concepts while complete supervision enables automatic concept discovery, properties useful for specific use-cases.}
    \label{fig:concept-sup}
\end{figure}

A visual summary of the two paradigms is provided in Figure~\ref{fig:concept-sup}. We demonstrate the two extreme ends of the spectrum - completely supervised and completely unsupervised concept learning. We utilize an example from \cite{koh2020concept}. With improved supervision, there is greater trust and control over the learned concepts. On the other hand, no supervision reduces trust but improves the efficiency of concept learning. All approaches discussed in this paper lie somewhere in the middle of this spectrum.

\section{Problem Setup}
\subsection*{Supervised Concept Learning}
 The supervised concept learning problem is characterized as a two-step process as detailed \cite{koh2020concept} for a given training sample $\{\mathbf{x},\mathbf{y},\mathbf{C}\}$. Given an observation $x$, a concept-based model learns a function $f$ to map the input to its associated human-understandable concepts $\mathbf{C} = \{c_1,\cdots,c_N\}$. Subsequently, a predictor $g$ projects the concept embeddings to $\hat{y}$, which is the prediction for the label $y$ of the sample in a downstream task. Mathematically $f,g$ are learned by minimizing the objective,
 \begin{equation}
     \arg \min_{f,g} \mathcal{L}_{Y} \left( g(f(\mathbf{x}), \mathbf{y} \right) + \lambda \mathcal{L}_{\mathbf{C}} \left( f(\mathbf{x}), \mathbf{C} \right)
 \end{equation}
 where $\mathcal{L}_{\mathbf{C}},\mathcal{L}_{\mathbf{Y}}$ are appropriate losses for concept learning and task learning respectively, while $\lambda$ is a hyperparameter.

\subsection*{Unsupervised Concept Learning}
The unsupervised concept learning problem is characterized as an end-to-end process as detailed in \cite{alvarez2018towards} for a given training sample $\{\mathbf{x},\mathbf{y}\}$. Given an observation $\mathbf{x},\mathbf{y}$, an encoder function and a relevance network $\mathbf{f,h}$, map from the input space to a concept representation space with dimensions $d$, while an aggregation function $a$ aggregates concept and score representations and projects them to the label $\hat{y}$. The encoder function and the relevance network output concept representations of the form $\{c_1,\cdots,c_N\}\in \mathcal{R}^d$ and relevance scores $\{s_1,\cdots,s_N\}\in \mathcal{R}^d$ associated with each concept respectively, while the aggregation function performs a weighted sum of the concept representations and the relevance scores. Mathematically, the relevance network $\mathbf{h}$ outputs a set of score vectors $\mathcal{S} =\{s_1, .. s_k\}$ for an input sample $x$. Mathematically,
\begin{equation}
     \arg \min_{f,h,a}~\mathcal{L}_{Y} \left(a(f(\mathbf{x})\odot~h(\mathbf{x})), \mathbf{y} \right)
\end{equation} 
where $\odot$ represents the element-wise product. In addition to the objective multiple models enforce strict regularizations on the learned concepts to ensure interpretability.

\section{Supervised Concept Learning}
\begin{figure*}[t]
\centering
\includegraphics[width=0.85\textwidth]{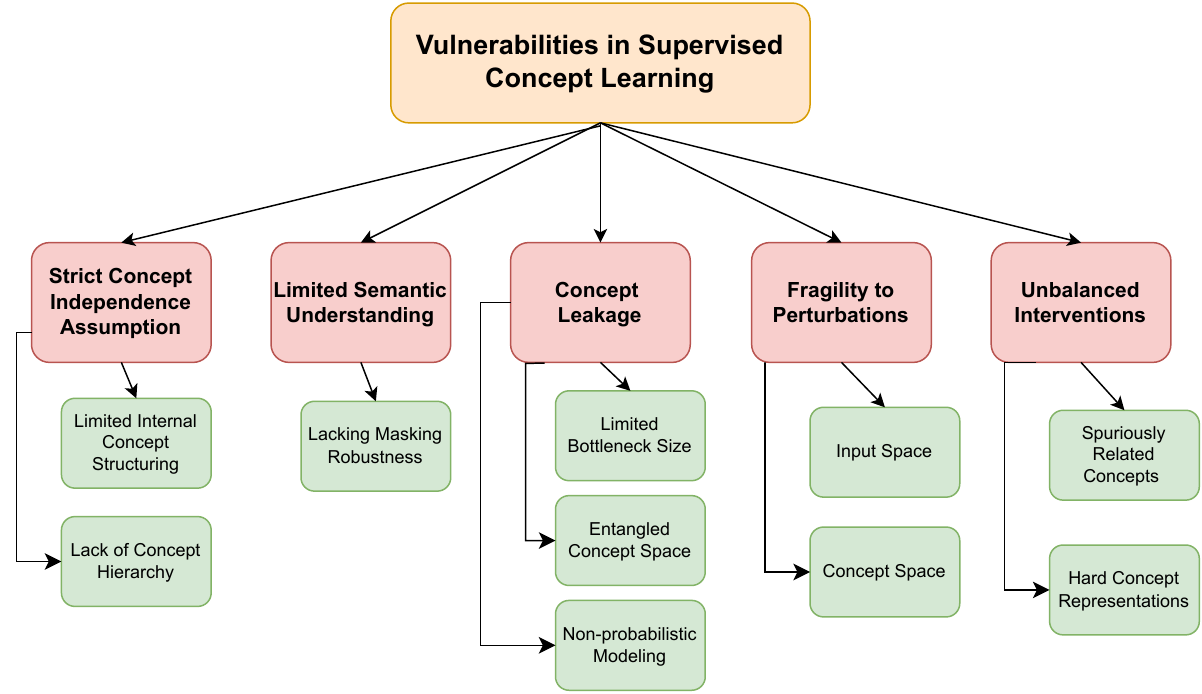}
\caption{Schematic figure illustrating key challenges in Supervised Concept Learning, highlighting five major vulnerabilities - Concept Leakage, Limited Semantic Understanding, Strict Concept Independence Assumption, Fragility to Perturbations, and Unbalanced Interventions (shown in red). Additionally, the generally postulated reasons for the identified vulnerabilities are listed below each vulnerability (shown in green). NOTE: The ordering has been modified from the original text for better readability.}
\label{fig:supervised-schematic}
\end{figure*}

In this section, we discuss the five most important limitations encountered by Concept-based Models in a supervised concept-learning setting - Concept Leakage, limited semantic understanding, strict concept independence assumption, fragility to perturbations, and unpredictable interventions. For each limitation, we further discuss the specific rationales for the observed behavior and studies that demonstrate these. We demonstrate a schematic of the discussed vulnerabilities in Figure~\ref{fig:supervised-schematic}. In addition, we list the strengths and weaknesses of supervised concept-based model architectures in Table~\ref{tab:vulnerability-table}. Finally in Table~\ref{tab:attack-research} provides an overview of the most important studies which systematically uncover the discussed vulnerabilities. 

\subsection{Concept Leakage}
The first major issue with CBMs \cite{koh2020concept} was identified as Concept Leakage. Concept Leakage refers to the model encoding non-relevant information in the concept labels i.e. extra information from the distribution itself. The problem was first identified in \cite{mahinpei2021promises} where they found that joint learning of concepts and tasks caused significant leakage even if the concepts were explicitly decorrelated. They also found that hard concept representations (fixing the labels to binary values) cause significantly more leakage as compared to soft representations where the values are not binary. As leakage is the most fundamental drawback of Supervised Concept-Learning, multiple approaches attempt to understand and mitigate it. Below we list the common reasons identified.

\begin{itemize}
    \item  \textbf{Limited Bottleneck Size:} CBMs with Additional Unsupervised Concepts \citep{sawada2022concept} postulated that the reason for leakage is the limited size of the concept bottleneck space as compared to the input image feature space which makes the bottleneck concepts encode extra non-concept relevant information. As a mitigation strategy, the authors propose increasing the bottleneck size by incorporating unsupervised concepts and learning them using unsupervised approaches like \cite{alvarez2018towards}. Similarly \cite{zabounidis2023benchmarking} utilizes both unsupervised concepts and a residual stream to fix leakage. Some studies outright remove or edit concept sets to improve predictive performance \cite{hu2024editable} during test time.
    
    \item \textbf{Entangled Concept Space:} A separate line of thought attempts to explicitly extract the concept-specific information before actually encoding it into the concepts. \cite{havasi2022addressing} postulated that entanglement in the concept space is the primary reason for leakage in CBMS. They propose to relax the Markovian assumptions behind concept learning where concepts are directly dependent on the data itself and tasks are dependent on concepts. They propose a `side-channel' wherein leakage is fixed by directly learning the task labels using both images and concepts rather than just concepts. Further, GlanceNets \citep{marconato2022glancenets} and CoLiDR \citep{sinha2024colidr} utilize a Variational Autoencoder (VAE) to first disentangle the image space into independent probability distributions dubbed `Generative Factors'. GlanceNets utilizes a framework that aligns concepts with a subset of Generative Factors implying supervision during disentanglement. Further, they utilize an open-set recognition (OSR) module to discard out-of-distribution samples to mitigate leakage.  CoLiDR on the other hand does not supervise the disentanglement procedure and further utilizes a stage-wise training procedure to map the Generative Factors to the concepts. Finally, another study \cite{bahadori2020debiasing} discusses the data and concept generation process from a causal perspective, which aids in debiasing CBMs and achieving more robust concept learning.
    
    \item  \textbf{Non-probabilistic Modeling of Concepts:} In most CBM approaches, the concepts are usually treated as fixed entities with binary labels while some studies model the data generation process probabilistically like \cite{marconato2022glancenets}. As a consequence, Concept Energy Models \citep{xu2024energy} was proposed - modeling inputs, concepts, and targets as outputs of neural networks and calculating the joint energy between them - reducing leakage and improving prediction performance. Similarly Probabilistic CBMs \citep{kim2023probabilistic} model the concepts as sampled from a probability distribution.
\end{itemize}

\subsection{Limited Semantic Understanding}
Ever since CBMs have been proposed, multiple studies have questioned whether they truly \textit{understand} the semantic content of the image. This circumspection is understandable, as there is no particular signal for the concept learning procedure to be grounded to specific areas in the image itself. \cite{margeloiu2021concept} conducted one of the first studies that systematically studies the concepts learned by CBMs by using saliency maps associated with each concept. They find that saliency maps do not just focus on the concept locations but also encode irrelevant entities like background noise. Future studies observe similar phenomena and propose solutions like \citep{huang2024concept}, which uses layer-specific prototypes to improve the grounding of concepts. Next, \cite{raman2024concept} designed a study involving the behavior of CBMs when a portion of the image containing the relevant concept is masked. They find that CBMs still output certain concept values for the concept wholly occluded by the masked section. This observation further demonstrates the lack of semantic understanding in CBMs. Some studies have attempted to understand and improve semantic understanding with Vision Language Models \citep{tan2024explain} and theoretical frameworks as proposed in \cite{crabbe2022concept}.

\subsection{Strict Concept Independence Assumption} 
CBMs work on the underlying assumption that concepts are mutually independent of each other. This assumption might be valid for datasets with visually identifiable grounded concepts (like color, and presence of attributes) but proves challenging for datasets with more semantic concepts (like shape). As a consequence, semantic concepts modeled as grounded concepts suffer from leakage and entanglement. An independent recent line of research attempts to address this problem by relaxing the independence criterion of concepts.

\begin{itemize}
    \item  \textbf{Internal Concept Structuring:} \cite{raman2024understanding} studies how well CBMs learn relationships between concepts and proposes an approach to structure these relationships from their internal embeddings during test-time.  For instance, \cite{sun2024eliminating} structure concepts into semantic hierarchies with an explicit intervention awareness matrix for improved interpretability. 
    \item \textbf{Concept Hierarchy Organization:} One of the first works to utilize a hierarchical structure to relax the concept independence criterion in CBMs was \cite{pittino2023hierarchical} and \cite{panousis2023hierarchical} which map fine and coarse-grained concepts hierarchically using a VLM. More recently, concept understanding has been modeled as decision trees \citep{ragkousis2024tree,liu2023primenet} to eliminate leakage, enhance explainability, and provide a neuro-symbolic pathway for prediction. \cite{sun2024eliminating} models the concept structure as a 2-level tree with structuring inputs generated by a Large Language Model (LLM). Lastly, PEEB \citep{pham2024peeb} uses a hierarchical structure and subsequent image grounding model to ground visually identifiable concepts to the image.
\end{itemize}

\begin{table*}[t]
    \centering
    \renewcommand{\arraystretch}{1.4} 
    \setlength{\tabcolsep}{10pt}  
    \resizebox{\textwidth}{!}{
    \begin{tabular}{c|ccc|ccc}
        \toprule
        \textbf{Model} & \textbf{Concept} & \textbf{Generative} & \textbf{Concept} & \textbf{Intervention} & \textbf{Semantic} & \textbf{Perturbation}\\
        \textbf{Architecture} & \textbf{Leakage} & \textbf{Process} & \textbf{Structuring} & \textbf{Awareness} & \textbf{Understanding} & \textbf{Robustness}\\
        \midrule
        CBMs \citep{koh2020concept}  & \xmark & \xmark & \xmark & \xmark & \xmark & \xmark  \\
        UnsupCBM \citep{sawada2022concept} & \xmark & \xmark & \xmark & \greencheck & \xmark & \xmark  \\
        CEM \citep{zarlenga2022concept} & \xmark & \xmark & \xmark & \greencheck & \xmark & \greencheck  \\
        GlanceNets \citep{marconato2022glancenets} & \greencheck & \greencheck & \xmark & \xmark & \greencheck & \xmark   \\
        ProbCBM \citep{kim2023probabilistic} & \xmark & \xmark & \xmark & \greencheck & \greencheck & \xmark  \\
        CEM (Energy) \citep{xu2024energy} & \xmark & \greencheck & \xmark & \xmark & \greencheck & \xmark\\
        InstCEM \citep{espinosa2024learning} & \greencheck & \xmark & \xmark & \greencheck & \greencheck & \greencheck\\
        CoLiDR \citep{sinha2024colidr}& \greencheck & \greencheck & \xmark & \xmark & \greencheck & \xmark\\
        Editable CBM \citep{hu2024editable} & \xmark & \xmark & \xmark & \greencheck & \xmark & \xmark  \\
        SupCBM \citep{sun2024eliminating} & \greencheck & \xmark & \greencheck & \greencheck & \xmark & \xmark  \\    
        \bottomrule
    \end{tabular}}
    \caption{Consolidated table listing concept-based model architectures proposed as an improvement over CBMs \cite{koh2020concept}. Note the architectures are sorted by recency of publication. We categorize all architectures based on two sets. The first set represents modifications in architectures especially designed to prevent vulnerabilities - concept leakage, modeling the generative process of data through probabilistic modeling, and presence of structuring in the concept space. The second set represents the properties which the model architectures should follow during inference-intervention awareness (implying balanced interventions), semantic understanding (implying robustness to masked/occlusion samples) and finally robustness to perturbations in the input/concept space. Note that none of the architectures satisfy all the 6 vulnerabilities - making future research in designing more effective model architectures an open problem.}
    \label{tab:vulnerability-table}
\end{table*}

\subsection{Fragility to Perturbations}
Robustness to perturbations is a well-studied problem in DNNs \cite{szegedy2013intriguing}. The primary reason for the existence of adversarial perturbations in DNNs is due to the high dimensional nature of the input space wherein small changes cause major changes in the predictions. As concept-based models learn dual objectives - where the target space is either predictions or concepts. Hence, the perturbations can be calculated both in the input space (with task and concept targets) and the concept space (with task targets).
\subsubsection{Perturbations in Input Space}
\cite{sinha2023understanding} first studied adversarial attacks on CBMs in the input space with concept and task labels as targets. Specifically, they demonstrate 3 types of attacks - erasure (targeting and removing a single concept), introduction (increasing influence of a specific concept) and confounding (removal and introduction of a set of concepts) in the concept space while keeping the prediction labels unchanged. In addition, they discuss a defense mechanism that smoothes the sparse concept space to defend against such attacks. As the concept space is even more sparse than the input space (albeit in smaller dimensions), the susceptibility to concept attacks is even more profound. To quantify the risk of such CBMs (Vulnerability), \cite{rasheed2024exploring} propose utilizing reduction in uncertainty, inspired by Mutual Information and can serve as a proxy for the threat level of CBMs. \cite{li2025factor} utilized a factor network to limit attack potential.

\subsubsection{Perturbations in Concept Space}
Concept-level Backdoor Attack (CAT) \citep{lai2024cat} proposes a modified version of backdoor attacks as first proposed in \cite{chen2017targeted}. CAT computes a standard concept subset to be substituted in the concept set to guide the CBMs to output a specific prediction. These concept subsets are referred to as triggers and are computed globally. Subsequently \cite{lai2024guarding} proposes a defense for backdoor attacks by proposing a majority vote by simpler classifiers during the inference phase to detect the poisoned concept.

\begin{table}[h]
    \centering
    \renewcommand{\arraystretch}{1.4} 
    \setlength{\tabcolsep}{10pt}  
    \resizebox{0.49\textwidth}{!}{
    \begin{tabular}{ccc}
        \toprule
        \textbf{Vulnerability} & \textbf{Proposed} & \textbf{Mitigation} \\
        \midrule
        Concept Leakage & \cite{margeloiu2021concept} & \xmark \\
                        & \cite{sawada2022concept} & \greencheck \\
        \hline
        Semantic Understanding & \cite{raman2024concept}  &  \xmark\\
        \hline
        Adversarial Robustness & \cite{sinha2023understanding}  & \greencheck  \\
        \hline
        Intervention Awareness & \cite{shin2023closer} & \xmark \\  
        & \cite{sheth2022learning} & \greencheck \\ 
        & \cite{steinmann2023learning} & \greencheck \\ 
        & \cite{chauhan2023interactive} & \greencheck \\ 
        
        \bottomrule
    \end{tabular}}
    \caption{In this table, we list the vulnerability studies performed on Concept-based models in a supervised setting. We report the vulnerability type in the first column, the works which conduct a systematic study on it in the second column and whether an effective mitigation strategy is proposed or not in the third column. Note that there has not been a systematic analysis of the inter concept structuring vulnerability - which remains an active area of research.}
    \label{tab:attack-research}
\end{table}

\subsection{Unbalanced Interventions}
One of the most appealing features of CBMs is that they are `human-in-the-loop' systems wherein human stakeholders can swiftly `correct' wrong predictions in case of misclassification by a CBM. 

\begin{itemize}
    \item \textbf{Spuriously-linked Concepts:} A significantly thorough study conducted by \cite{shin2023closer} performed systematic interventions in large-scale CBMs. Through a series of concept selection strategies, they observe a significant flaw wherein some intervened concept subsets do not `fix' the wrong prediction, and others fix it easily, as measured by budget and changes in performance. This observation is also found across certain concepts with null values, implying that intervention is a direct consequence of the training procedure and cannot be assumed to be an independent and reliable entity due to the spuriously learned concept relationships. Other studies such as \cite{sheth2022learning} utilize uncertainty estimations to select concepts to intervene and achieve maximal prediction performance. Finally, studies such as \cite{chauhan2023interactive} learn intervention policies using Reinforcement Learning to improve prediction performance post-intervention. Finally, studies like \cite{singhi2024improving} realign concept representation to be mutually de-correlated to achieve better concept interventions and maintain high predictive performance while studies like \cite{steinmann2023learning} utilize a intervention memory to perform future interventions well.
    
    \item \textbf{Hard Concept Intervention Problem:} A few studies recognize the fact that human stakeholders cannot perform interventions on soft concept representations and only `fix' binary concept labels. To alleviate this, Concept Embedding Models \citep{zarlenga2022concept} were proposed which consist of both a `presence' and `absence' attribute in each concept representation with binary values. As a result, the model learns from counterfactual concept representations as well, further smoothening the concept space. A further iteration of CEMs - IntCEM \citep{espinosa2024learning} was proposed which adds intervention awareness in CEMs. IntCEMs identify concepts which maximize performance after interventions.

\end{itemize}

\section{Unsupervised Concept Learning}
\begin{figure}[h]
    \centering
    \includegraphics[width=0.48\textwidth]{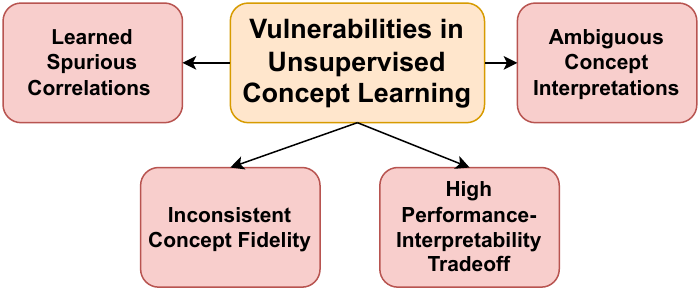}
    \caption{Schematic figure illustrating key challenges in Unsupervised Concept Learning, highlighting four major vulnerabilities - Learned Spurious Correlations, Ambiguous Concept Interpretations, Inconsistent Concept Fidelity, and High Performance-Interpretability Tradeoff (shown in red).}
    \label{fig:unsupervised-schematic}
\end{figure}

In this section, we discuss the four most important limitations of Concept-based Models trained under an unsupervised setting, i.e. concept discovery is guided by task learning without any human-annotated concepts. These include - spuriously learned concepts, ambiguous concept interpretation, reduced concept fidelity, and a very high performance-interpretability tradeoff. Figure~\ref{fig:unsupervised-schematic} gives a schematic overview of the discussed vulnerabilities while Table~\ref{tab:senn-vulnerable}

\begin{table*}[t]
    \centering
    \renewcommand{\arraystretch}{1.4} 
    \setlength{\tabcolsep}{10pt}  
    \resizebox{0.8\textwidth}{!}{
    \begin{tabular}{c|cc}
        \toprule
        \textbf{Vulnerability} & \textbf{Proposed} & \textbf{Mitigation Strategy} \\
        \midrule
        Learned Spurious  & \cite{sawada2022c} & Contrastive Training + Object detectors \\
        Correlations     & \cite{ahmad2024causal} & Causal relationships + Graph Structure  \\
        & \cite{disenn} & Disentangled representation Learning  \\
        \hline
        Ambiguous Concept & \cite{wang2023learning}   &  Slot Attention + Self-supervised Learning \\
        Interpretations  & \cite{norrenbrock2024q}   &  Concept Selection (Quantization) + Local explanations\\
                         & \cite{teso2019toward} &    Local explanations    \\
        \hline
        Inconsistent Concept & \cite{sinha2024selfexplain}  & Prototype based iterative grounding + Self Supervised Learning  \\
        Fidelity &   & \\
        \hline
        High Performance-   & \cite{norrenbrock2024q} & Side Channels + Concept Selection  \\  
        Interpretability Tradeoffs & \cite{sarkar2022framework} & Side Channel + Self supervised Learning \\ 

        \bottomrule
    \end{tabular}}
    \caption{Consolidated table listing the common vulnerabilities observed in unsupervised concept learning. We categorize all the discussed methods into four categories of vulnerabilities (Column-1) and list the mitigation strategies employed in Column-3. We observe that all approaches attempt to improve on unsupervised concept learning using a variation of self-supervised or contrastive learning. We believe that further research is required in these domains wherein the concept representations can be accurately inferred and understood.}
    \label{tab:senn-vulnerable}
\end{table*}

As opposed to supervised concept learning, unsupervised approaches are less studied as there is still a high reluctance to automatically discover concepts (as opposed to labeled and immutable concepts) even though they offer more scalability and flexibility.

\subsection{Learned Spurious Correlations}
Unlike supervised settings, unsupervised concept learning is a more challenging setting. Note that concepts being flexibly learned make them highly dependent on the task as opposed to fixed and immutable as in the case of supervised settings. This makes concept learning amplify common problems encountered with standard model training - such as spurious correlations learned through dataset bias. 
Contrastive-SENN (C-SENN), as proposed in \cite{sawada2022c} demonstrated the problem of spurious concept correlation learning through a dash camera-based dataset wherein the background features dominate while more important concepts such as red lights, and stop-lights are ignored. As a consequence, they propose contrastive learning to impart weak supervision in the concept learning procedure which is able to mitigate the problem to an extent. In a parallel study, owing to the dominance of spuriously learned attributes from dataset bias, \cite{disenn} proposes to disentangle the input data using a Variational Autoencoder (VAE) and subsequently utilize the learned probability distributions as priors to concept learning - providing a clear and transparent data generation mechanism. Finally, this observation is echoed in \cite{ahmad2024causal}, where commonly used unsupervised concept learning methods suffer from spurious correlations which is mitigated by forming pre-defined causal graphs.

\subsection{Ambiguous Concept Interpretations}
One of the most fundamental challenges in unsupervised concept learning is the mechanism to interpret concepts. As there is no straightforward way to interpret representations, methods like \cite{alvarez2018towards} utilize concept activations as \textit{proxies} for importance, and the actual interpretation is done through prototypes from the training set. This creates multiple challenges as the semantic understanding of the set of explanation prototypes is left to the stakeholder. To alleviate these issues BotCL \citep{wang2023learning} was proposed to provide local and grounded explanations through saliency maps in addition to prototypes using slot attention. Parallely, Q-SENN \citep{norrenbrock2024q} utilizes a feature selection `quantization' layer to only select the most important concepts responsible for predictions. This in turn, provides both global explanations (from prototypes) and local explanations (from saliency maps) adding trust to only global explanations. Yet another study by \cite{teso2019toward} utilizes SENN as an aid to improve local explanations (Note: opposite to Q-SENN) and provides a more holistic explanation landscape. To sum up, we observe that interpreting concept representations is a challenging problem and still requires significant research. Lastly, the study by \cite{qian2024towards} utilizes a conformal prediction wherein the concept uncertainty is utilized as an aid for prediction and interpretations.

\subsection{Inconsistent Concept Fidelity}
A byproduct of utilizing task labels to guide concept learning is that the concepts learned are extremely useful for solving the task rather than being explainable. This behavior manifests itself in reducing \textit{fidelity} - a fundamental aspect of concepts that entails samples from the same class sharing a majority of concepts. However, \cite{sinha2024selfexplain} observed that concepts learned by unsupervised methods are often completely different, even though the task performance is comparable to non-explainable models. This is further exacerbated by comparing concept fidelity \textit{across} domains with the same task. As a mitigation strategy, \cite{sinha2024selfexplain} utilized a prototype-based grounding approach to better ground the concepts in a single domain and reduce concept drift during training.

\subsection{High Performance-Interpretability Tradeoffs}
Finally, unsupervised concept-learning models are especially susceptible to high performance-interpretability tradeoffs as compared to other approaches. This observation can be seen in multiple works \cite{alvarez2018towards, wang2023learning,norrenbrock2024q}. This behavior mainly occurs due to a high amount of regularization on concepts to ensure interpretability and no additional information signals as in supervised learning methods. For example, consider SENN\cite{alvarez2018towards}. The regularization imposed on the concepts - the robustness loss includes a Jacobian calculation of the concept encoder with respect to the inputs. This makes the concept space heavily regularized - making learned concepts more grounded (at in essence more interpretable) but reducing the task performance. Some works have attempted to use a `side-channel' approach to directly map the inputs to the tasks, with concept learning as a secondary objective to task performance. Although this increases task performance, studies still need to be conducted on their effects on interpretability. The study by \cite{sarkar2022framework} was one of the first works utilizing the `side-channel' approach and has been used extensively by others such as \cite{sinha2024selfexplain} to attain state-of-the-art prediction performances.

\section{Future Research}
\label{sec:future}
\subsection{LLM/VLM based Concept Discovery}
Even though some approaches like \cite{pham2024peeb} and \cite{chauhan2023interactive} utilize Large Language Models and Vision Language Models to improve concept structuring and interventions respectively, a large future direction of research utilizes these models as concept extracting and discovery \textit{agents}. In addition, LLM linguistic reasoning combined with VLM localization abilities can help improve concept structure, learn causal relationships, and improve concept interactions. Using an LLM's encoded knowledge can act as an ontology for concept structuring - aiding discovery without manual annotations. Similarly, creative prompting methods can extract representative concepts for training concept-based models. Finally, embeddings from VLMs can act as concept priors in unsupervised concept-based model training to guide both concept and task discovery. Lastly, utilizing generative models for generating concepts in a counterfactual approach is not well-studied.

\subsection{Privacy Issues in Concept-based Models}
With increasing adoption of CBMs in sensitive applications, apart from the aforementioned vulnerabilities, privacy-related vulnerabilities need to be solved as well. For example, DNNs have been shown to be susceptible to Membership Inference Attacks, a well-known vulnerability in sensitive domains like medicine and finance. CBMs strive to be widely adopted in similar sensitive domains and should be evaluated against standard privacy-preserving attacks. Further, improving Concept-based Model's defenses to such attacks by changes in architecture, training procedure and data curation process is a promising area of research including Differential Privacy, privacy protection training, etc.

\section{Conclusion}
Concept-Based Models have quickly emerged as go to inherently explainable models as they incorporate human-understandable concepts into decision-making. However, as highlighted in this survey, supervised concept learning models face several critical limitations that must be addressed for their widespread adoption in high-stakes applications. Supervised models suffer from issues such as concept leakage, limited semantic understanding, strict independence assumptions, fragility to perturbations and unbalanced interventions. We also observe that designing better architectures is an important area of ongoing research. Next, unsupervised concept learning models are vulnerable to spuriously learned concepts, inconsistent concept fidelity, high performance-interpretability tradeoffs, and ambiguous concept interpretations. Several mitigation strategies have attempted to reduce these vulnerabilities by mainly employing creative training mechanisms like self-supervised learning. We hope that this survey aides and improves research in this topic.

\bibliographystyle{named}
\bibliography{ijcai25}

\end{document}